%% file: neurips_2024.tex
\documentclass{article}

     \usepackage[preprint]{neurips_2024}

\usepackage[utf8]{inputenc} 
\usepackage[T1]{fontenc}    
\usepackage{hyperref}       
\usepackage{url}            
\usepackage{booktabs}       
\usepackage{amsfonts}       
\usepackage{nicefrac}       
\usepackage{microtype}      
\usepackage{xcolor}         

\usepackage{pgfplots}
\pgfplotsset{compat=1.18}
\usepackage{amsmath}
\usepackage{subcaption}
\usepackage{makecell}
\usepackage{float}

\usetikzlibrary{external}
\title{QuAILoRA: Quantization-Aware Initialization for LoRA}

\author{Neal Lawton \\
\texttt{lawtneal@amazon.com} \\
\And
Aishwarya Padmakumar \\
\texttt{padmakua@amazon.com} \\
\And
Judith Gaspers \\
\texttt{gaspers@amazon.com} \\
\And
Jack FitzGerald \\
\texttt{jgmf@amazon.com} \\
\And
Anoop Kumar \\
\texttt{anooamzn@amazon.com} \\
\And
Greg Ver Steeg \\
\texttt{gssteeg@amazon.com} \\
\And
Aram Galstyan \\
\texttt{argalsty@amazon.com}
}

\begin{document}

\maketitle

\begin{abstract}
\input{abstract}
\end{abstract}

\section{Introduction}
\input{intro}

\section{Related Work}
\input{related_work}

\section{Method}
\input{method}

\section{Experiments}
\input{experiments}

\section{Conclusion}
\input{conclusion}

\section{Limitations}
\input{limitations}

\bibliography{neurips_2024}
\bibliographystyle{abbrvnat}

\newpage


\appendix

\section{Appendix}

\input{appendix}

\end{document}

%% file: abstract.tex
QLoRA reduces the memory-cost of fine-tuning a large language model (LLM) with LoRA by quantizing the base LLM. However, quantization introduces quantization errors that negatively impact model performance after fine-tuning. In this paper we introduce QuAILoRA, a quantization-aware initialization for LoRA that mitigates this negative impact by decreasing quantization errors at initialization. Our method spends a small amount of computational overhead to compute this quantization-aware initialization, without increasing the memory-cost of fine-tuning. We evaluate our method on several causal language modeling and downstream evaluation tasks using several different model sizes and families. We observe that almost all LLMs fined-tuned with QuAILoRA achieve better validation perplexity. When evaluated on downstream tasks, we find that QuAILoRA yields improvements proportional to the negative effect of quantization error. On average, applying QuAILoRA to 4-bit QLoRA models yields 75\% of the validation perplexity decrease and 86\% of the downstream task accuracy increase as doubling the quantization precision to 8-bit, without increasing GPU memory utilization during fine-tuning.



%% file: intro.tex
Fine-tuning state-of-the-art large language models (LLMs) requires a large amount of computational resources due to their increasingly large size. 
QLoRA \citep{dettmers2023qlora}, a quantized version of LoRA \citep{hu2021lora}, reduces the memory-cost of fine-tuning sufficiently to fine-tune LLMs on the order of 70B parameters on a single GPU, making fine-tuning much more convenient and accessible. Although quantization greatly reduces memory costs, it also introduces quantization errors that negatively impact the task performance of the model after fine-tuning. In this paper we propose Quantization-Aware Initialization for LoRA (QuAILoRA), a method for initializing the LoRA matrices of a QLoRA model to reduce quantization errors. When fine-tuning a model with QLoRA, each parameter matrix of the fine-tuned model takes the form $Q + AB^\top$, where $Q$ is the quantization of the base parameter matrix $W$ and $AB^\top$ is the low-rank LoRA update. Typically the matrix $A$ is initialized random normal and the matrix $B$ is initialized zero so that the input-output mapping of the QLoRA model is the same as the quantized base model at initialization. Instead, we propose spending a small amount of computational overhead to find an initialization for the LoRA matrices so that the input-output mapping of the QLoRA model is more similar to the full-precision base model at initialization.

We conduct an extensive set of experiments and establish that QuAILoRA (1) is robust to the choice of calibration set; (2) yields better validation perplexity than the baseline initialization across many model families and sizes on several causal LM tasks; (3) yields consistently positive results on downstream task evaluations for smaller, lower-precision quantized LLaMA models. 
Additionally, we establish that our method is increasingly effective with larger LoRA ranks and does not appear to affect the rate of convergence of fine-tuning compared to the baseline initialization. QuAILoRA provides the largest benefit when the negative effect of quantization error is significant. 

%% file: related_work.tex
Our method improves the performance of QLoRA, a parameter-efficient fine-tuning (PEFT) technique, but  many other PEFT methods exist in the literature, including adapter-based strategies \citep{houlsby2019parameter}, BitFit \citep{zaken2021bitfit}, diff-pruning \citep{guo2020parameter}, NAS for PEFT \citep{lawton2023neural}, and AdaLoRA \citep{zhang2023adaptive}. These methods and others can be combined to form PEFT design spaces \citep{he2021towards}. 

We propose a method for computing a quantization-aware initialization of the trainable parameters of a QLoRA model \citep{hu2021lora, dettmers2023qlora}. Our method exploits the special low-rank structure of the updates to efficiently compute such an initialization. However, it is possible our method could be extended to other reparamterization-based PEFT strategies that exist in the literature, such as Kroncker-product fine-tuning updates \citep{he2022parameter}. 

We seek a quantization-aware initialization of the trainable LoRA parameters of a QLoRA model that reduces quantization errors between the QLoRA model and the full-precision model. In order to do so, we build off techniques from the literature on post-training quantization, such as GPT-Q \citep{frantar2022gptq}, OPS \citep{frantar2022optimal}, Bit-stitching \citep{wang2020towards}, and QuIP \citep{chee2023quip}. Like GPT-Q, we optimize a calibrated quantization objective to decrease quantization errors on a target calibration dataset. Our method is similar in strategy to other recent methods in the literature such as LQ-LoRA \citep{guo2023lq}, LoftQ \citep{li2023loftq}, and ApiQ \citep{liao2024apiq}.



%% file: method.tex
\subsection{Background and notation}

LoRA \citep{hu2021lora} is a parameter-efficient fine-tuning method that fine-tunes a model by learning a low-rank update $AB^\top$ for each parameter matrix $W$, where $W$ is an $m \times n$ matrix, $A$ is an $m \times r$ matrix, and $B$ is an $n \times r$ matrix, where $r$ is the user-specified LoRA rank. After fine-tuning, the parameter matrix in the fine-tuned LoRA model is $W + AB^\top$. In contrast, QLoRA \citep{dettmers2023qlora} fine-tunes a quantized version of the model with LoRA so that the parameter matrix in the fine-tuned QLoRA model is $Q + AB^\top$, where $Q$ is the quantization of $W$.

Typically, $A$ is initialized random normal and $B$ is initialized zero. We refer to this as the "baseline initialization".

\subsection{Calibrated quantization}

To find a quantization-aware initialization for the LoRA matrices $A$ and $B$, we minimize a \textit{calibrated quantization} objective that aims to keep the activations of the QLoRA model close to those of the full-precision base model on a given \textit{calibration dataset} at initialization. For each parameter matrix $W$, we propose initializing the LoRA matrices $A$ and $B$ by minimizing a calibrated quantization objective, defined

\begin{equation}
\min_{A,B} \frac{1}{2} \|(W - (Q + AB^\top))X\|_F^2,
\end{equation}

where $Q$ is the quantization of $W$, $A$ and $B$ are the real (stored in high-precision) LoRA matrices, $\|\cdot\|_F$ is the Frobenius norm, and $X$ is an $n \times s$ real (stored in high-precision) matrix consisting of the input activations of the full-precision base model to the parameter matrix $W$ on a calibration dataset of $s$ examples. We minimize this objective with respect to $A$ and $B$ independently for each parameter matrix $W$ in the model before proceeding to fine-tune as usual with QLoRA.

The samples used for the calibration dataset may come from the training data of the task that we plan to fine-tune on or from another source.

\subsection{Uncalibrated quantization}

A related objective that we will make repeated reference to is the \textit{uncalibrated quantization} objective, which does not use a calibration dataset or matrix of input activations $X$ and which instead aims to keep the weights, rather than the activations, of the QLoRA model close to those of the full-precision base model at initialization. The uncalibrated quantization objective is defined

\begin{equation}
\min_{A,B} \frac{1}{2} \|(W - (Q + AB^\top))\|_F^2.
\end{equation}

Note that if the input activations for the parameter matrix $W$ are uncorrelated, so that ${X X^\top = c \cdot I}$ for some scalar $c$, then optimizing the calibrated and uncalibrated quantization objectives are equivalent. However, we find that initializing $A$ and $B$ to minimize this uncalibrated quantization objective is ineffective for improving the performance of QLoRA over the baseline initialization. 


\subsection{Optimization}

To minimize the calibrated quantization objective with respect to $A$ and $B$, we propose a simple alternating optimization algorithm. To begin the optimization, we initialize $A$ and $B$ by minimizing the uncalibrated quantization objective. This optimization problem is solved by computing the SVD of the parameter quantization error $W - Q$:

\begin{equation}
W - Q = U \Sigma V^\top.
\end{equation}

Let $\Sigma_r$ be the diagonal matrix consisting of the $r$ largest singular values of $W-Q$, and let $U_r$ and $V_r$ be matrices consisting of the corresponding top left and right singular vectors, respectively. Then we initialize

\begin{align}
    A = U_r \sqrt{\Sigma_r} \qquad B = V_r \sqrt{\Sigma_r}.
\end{align}

After initializing $A$ and $B$, our algorithm proceeds to optimize the calibrated quantization objective alternately over $A$ and $B$. Define the activation correlation matrix $H := XX^\top$. Then the updates for $A$ and $B$ each involve solving an $r \times r$ linear system:

\begin{align}
    A &:= (W-Q) H B^\top (B H B^\top)^{-1} \\
    B &:= (A^\top A)^{-1} A^\top (W-Q)
    \label{eq:uv_updates}
\end{align}

Since $r$ is small, typically on the order of $64$, these linear systems are computationally inexpensive to solve. Note that to compute these updates, we only need to compute and store the $n \times n$ activation correlation matrix $H$ rather than the $n \times s$ matrix of input activations $X$, which is typically much larger.




In all our experiments with our method we use $s = 2000$ calibration samples (similar to \cite{frantar2022gptq}) and execute $20$ steps of alternating optimization, so that $A$ and $B$ are each updated $20$ times.

%% file: experiments.tex
We compare our method against a QLoRA baseline, quantized to either 4-bit or 8-bit precision, that uses the baseline initialization: $A$ random normal and $B$ zero. Whenever we fine-tune a QLoRA model, we use learning rate $2 \times 10^{-4}$ and fine-tune for one epoch using total batch size of 16. We perform quantization using the BitsAndBytes library \citep{dettmers2022llmint8, dettmers2022optimizers}, using double quantization (quantization of the affine quantization scaling constants) and the NormalFloat4 (NF4) data type for 4-bit quantization. This quantization configuration makes quantization errors small, even before applying our method. In all experiments, we use LoRA $\alpha=16$, gradient accumulation, warm-up ratio $0.03$,  and optimize using the AdamW optimizer during fine-tuning.

We evaluate our method on publicly available LLMs across four different LLM families: LLaMA \citep{touvron2023llama}, OPT \citep{zhang2022opt}, BLOOM \citep{scao2022bloom}, and Pythia \citep{biderman2023pythia}. We use publicly available causal language modeling datasets for calibration, training, and evaluation: Alpaca \citep{alpaca}, Unified-Chip2 \citep{chip2}, Self-Instruct \citep{wang2022self}, HH-RLHF \citep{bai2022training}, and SlimOrca \citep{SlimOrca}. Unless stated explicitly otherwise, we use LoRA rank $r=64$ and fine-tune for 1000 steps, except for Pythia-70m, which we fine-tune for 10k steps.

\subsection{Choice of calibration set}

\input{fig/calibration_choice_table}

First we evaluate the effect of the choice of the calibration dataset on  performance after fine-tuning. In each experiment, we calibrate and/or fine-tune on Alpaca, Unified-Chip2, Self-Instruct, or HH-RLHF. For each choice of calibration dataset and fine-tuning dataset, we report the validation perplexity after fine-tuning, averaged over six LLMs: Pythia-12b, Pythia-410m, Pythia-70m, BLOOM-3b, BLOOM-560m, and LLaMa-13b.

The results are in Table \ref{table:calibration_choice}. For comparison, for each fine-tuning dataset we include the average validation perplexity after fine-tuning for the 4-bit and 8-bit baseline QLoRA models. We observe that the choice of calibration dataset does not significantly affect task performance after fine-tuning: the difference in performance between our method and the 4-bit and 8-bit QLoRA baselines for any calibration dataset choice is much larger than the difference between the performance of our method with different calibration dataset choices. We conclude that our method is robust to the choice of calibration dataset.

\subsection{Perplexity after fine-tuning}

\input{fig/summarized_perplexity_table}

Here we compare the validation perplexity of 4-bit and 8-bit QLoRA models initialized with our method versus the baseline initialization after fine-tuning. Each QuAILoRA model in this section is calibrated on Alpaca.

The average validation perplexity after fine-tuning on Alpaca, Unified-Chip2, Self-Instruct, or HH-RLHF is in Table \ref{table:summarized_perplexity_table}, and a breakdown of this average by task is in Appendix Table \ref{table:full_perplexity_table}. Results for the 8-bit OPT models are omitted due to errors encountered while fine-tuning these models using the BitsAndBytes and \verb|peft| libraries. We observe that in the vast majority of cases, fine-tuning a 4-bit or 8-bit QLoRA model from our initialization achieves lower validation perplexity than fine-tuning from the baseline initialization. In a small number of cases, the model initialized with our method achieves worse validation perplexity compared to the baseline initialization, which we present as failure cases.

In most cases, the 4-bit QuAILoRA model outperforms the 4-bit QLoRA baseline and underperforms the 8-bit QLoRA baseline. We can use the results in Table \ref{table:summarized_perplexity_table} to compare the decrease in validation perplexity yielded by applying our method to a 4-bit QLoRA model versus doubling the quantization precision of the 4-bit QLoRA model to 8-bit. For each model, we can compute the gap in average validation perplexity closed by our method as the difference in average validation perplexity between the 4-bit QLoRA and QuAILoRA models, divided by the difference in average validation perplexity between the 4-bit and 8-bit QLoRA models, capping the computed ratio at $1$. The results of this analysis are in Table \ref{table:perplexity_gap_closed}. We exclude LLaMA-30b from this analysis, since we observed the 8-bit baseline model underperform the 4-bit baseline on average for this model. We also exclude the OPT-13b and OPT-30b LLMs since we could not generate results for the 8-bit versions of these models. Averaging across the other models, the average gap in perplexity between 4-bit and 8-bit quantization closed by applying our method to the 4-bit QLoRA models is $75\%$. We conclude that applying our method to 4-bit quantized QLoRA models yields approximately $75\%$ of the decrease in validation perplexity achieved by doubling the quantization precision, without increasing GPU memory utilization during fine-tuning.

We observe that in the LLaMA family of models, quantization error does not appear to significantly negatively affect validation perplexity after fine-tuning: the difference in validation perplexity after fine-tuning between the 4-bit and 8-bit baseline QLoRA models is small. Since our method improves performance by reducing quantization error, we observe that the gain in performance provided by our method over the baselines for the LLaMA models is proportionately small. For the other families, the difference in performance between the 4-bit and 8-bit baselines is significant enough for our method to provide a larger advantage over the baselines.

From these results, we conclude that QuAILoRA reduces validation perplexity after fine-tuning on average, proportional to the negative affect of quantization error.

\subsection{Performance on downstream tasks}

\input{fig/downstream_tasks_train_split_summarized}

Here we compare how LLaMA models fine-tuned with QuAILoRA versus QLoRA on Alpaca or SlimOrca perform on seven downstream tasks: Arc-Challenge (Arc-C) \citep{allenai:arc}, Arc-Easy (Arc-E), BoolQ \citep{clark2019boolq}, HellaSwag (HS) \citep{zellers2019hellaswag}, OpenBookQA (OBQA) \citep{OpenBookQA2018}, PIQA \citep{Bisk2020}, and WinoGrande (WinoG) \citep{ai2:winogrande}. We use the EleutherAI LM Evaluation Harness \citep{eval-harness} for evaluation. For Alpaca experiments, we calibrate on Alpaca and fine-tune for one epoch. For SlimOrca experiments, we calibrate on SlimOrca and fine-tune on a random size-10000 subset of SlimOrca. 

The average accuracy achieved across the evaluation tasks is in Tables \ref{table:downstream_tasks_alpaca_summarized} and \ref{table:downstream_tasks_slimorca_summarized}, and a breakdown of this average by task is in Appendix Table \ref{table:downstream_tasks_train_split_full}. The gap in accuracy between 4-bit and 8-bit quantization closed by applying our method to each 4-bit model is computed in Tables \ref{table:downstream_tasks_alpaca_gap_closed} and \ref{table:downstream_tasks_slimorca_gap_closed}. We compute the gap closed for the downstream task experiments in the same way as for the perplexity experiments in the previous subsection. Averaged over all the downstream task experiments, the average gap closed by our method is approximately 86\%. We conclude that our method improves the downstream task performance of QLoRA on average, and that applying our method to 4-bit quantized QLoRA models yields approximately 86\% of the increase in downstream task performance achieved by doubling the quantization precision, without increasing GPU memory utilization during fine-tuning. 









\subsection{Effect of LoRA rank}

\input{fig/lora_rank_fig}

Here we examine the effect of the choice of the LoRA rank hyperparameter $r$ on performance after fine-tuning. We expect that using larger $r$ will allow our initialization of $A$ and $B$ to reduce a greater part of the quantization error and result in better performance after fine-tuning. A plot illustrating the effect of changing the LoRA rank when using our initialization versus the baseline initialization, averaged across six 4-bit LLMs (excluding Pythia-70m, LLaMa-30b and OPT-30b) and 4 causal language modeling tasks, is in Figure \ref{fig:lora_rank}. We observe that the performance of QLoRA generally increases as we increase the LoRA rank $r$, albeit with diminishing returns. In contrast, the choice of $r$ does not appear to significantly affect the performance of QLoRA when using the baseline initialization.

\subsection{Convergence of fine-tuning}

\input{fig/convergence_plot}

Here we examine how our method affects the speed of convergence of fine-tuning. In Figure \ref{fig:convergence_plots}, we plot the convergence curve for each of the fine-tuning experiments used to generate the validation perplexity results in Table \ref{table:summarized_perplexity_table}. We plot fine-tuning steps on the horizontal axis and the average validation perplexity across the four causal language modeling tasks, measured every 100 fine-tuning steps, on the vertical axis. We observe that the 4-bit QLoRA models, when initialized with our method, achieve lower average validation perplexity at all stages of fine-tuning compared to the 4-bit QLoRA baselines. The difference in performance between the 8-bit QLoRA models fine-tuned with and without our method is on average much smaller compared to the 4-bit models, likely because the quantization error is already quite small for the 8-bit models and there are diminishing returns for reducing quantization error. From these plots, it appears that fine-tuning does not converge more quickly or slowly when initialized with our method compared to the baseline initialization. Rather, the benefit of decreased validation perplexity after fine-tuning observed for our method is due to decreased validation perplexity at initialization from decreased quantization error.

%% file: fig/calibration_choice_table.tex
\begin{table}[!t]
\centering
\begin{tabular}{|c|c|c|c|c|c|}
\hline
\makecell{ Fine-tuned $\rightarrow$ \\ Calibrated $\downarrow$} & alpaca & chip2 & s-i & rlhf \\
\hline
alpaca & 6.94 & 6.04 & 3.58 & 8.36 \\
chip2 & 6.95 & 6.04 & 3.56 & 8.35 \\
s-i & 6.94 & 6.04 & 3.58 & 8.36 \\
rlhf & 6.95 & 6.05 & 3.58 & 8.35 \\
\hline
QLoRA 4-bit & 7.02 & 6.13 & 3.65 & 8.43 \\
QLoRA 8-bit & 6.87 & 6.01 & 3.57 & 8.35 \\
\hline
\end{tabular}
\caption{Affect of calibration dataset choice on validation perplexity after fine-tuning for 4-bit models, averaged across six LLMs. We observe that the choice of calibration dataset does not significantly affect validation perplexity after fine-tuning.}
\label{table:calibration_choice}
\end{table}

%% file: fig/summarized_perplexity_table.tex
\begin{table*}[!t]
\caption{Validation perplexity results}
\label{table:summarized_perplexity_fig}
\begin{subtable}{.6 \linewidth}
\centering
\begin{tabular}{|cccc|}
\hline
Model & Bits & Method & Avg.\\
\hline
LLaMA-7b & 4 & QLoRA & $3.51$ \\
LLaMA-7b & 4 & Ours & $\bf 3.49$ \\
\hline
LLaMA-7b & 8 & QLoRA & $3.49$ \\
LLaMA-7b & 8 & Ours & $\bf 3.48$ \\
\hline
LLaMA-13b & 4 & QLoRA & $3.33$ \\
LLaMA-13b & 4 & Ours & $\bf 3.32$ \\
\hline
LLaMA-13b & 8 & QLoRA & $3.32$ \\
LLaMA-13b & 8 & Ours & $\bf 3.31$ \\
\hline
LLaMA-30b & 4 & QLoRA & $\bf 3.30$ \\
LLaMA-30b & 4 & Ours & $\bf 3.30$ \\
\hline
LLaMA-30b & 8 & QLoRA & $3.31$ \\
LLaMA-30b & 8 & Ours & $\bf 2.29$ \\
\hline
OPT-13b & 4 & QLoRA & $3.77$ \\
OPT-13b & 4 & Ours & $\bf 3.71$ \\
\hline
OPT-30b & 4 & QLoRA & $3.66$ \\
OPT-30b & 4 & Ours & $\bf 3.60$ \\
\hline
BLOOM-560m & 4 & QLoRA & $6.84$ \\
BLOOM-560m & 4 & Ours & $\bf 6.73$ \\
\hline
BLOOM-560m & 8 & QLoRA & $\bf 6.73$ \\
BLOOM-560m & 8 & Ours & $6.76$ \\
\hline 
BLOOM-3b & 4 & QLoRA & $4.82$ \\
BLOOM-3b & 4 & Ours & $\bf 4.75$ \\
\hline
BLOOM-3b & 8 & QLoRA & $4.78$ \\
BLOOM-3b & 8 & Ours & $\bf 4.76$ \\
\hline
Pythia-70m & 4 & QLoRA & $10.98$ \\
Pythia-70m & 4 & Ours & $\bf 10.80$ \\
\hline
Pythia-70m & 8 & QLoRA & $10.72$ \\
Pythia-70m & 8 & Ours & $\bf 10.69$ \\
\hline
Pythia-410m & 4 & QLoRA & $6.73$ \\
Pythia-410m & 4 & Ours & $\bf 6.67$ \\
\hline
Pythia-410m & 8 & QLoRA & $6.57$ \\
Pythia-410m & 8 & Ours & $\bf 6.54$ \\
\hline
Pythia-12b & 4 & QLoRA & $5.14$ \\
Pythia-12b & 4 & Ours & $\bf 5.11$ \\
\hline
Pythia-12b & 8 & QLoRA & $5.09$ \\
Pythia-12b & 8 & Ours & $\bf 5.08$ \\
\hline
\end{tabular}
\caption{Validation perplexity after fine-tuning various LLMs on four causal language modeling tasks, with and without QuAILoRA. Our method provides a moderate improvement over the baseline in the
vast majority of cases.}
\label{table:summarized_perplexity_table}
\end{subtable}
\begin{subtable}{.4 \linewidth}
\centering
\begin{tabular}{|cc|}
\hline
Model & Gap Closed \\
\hline
LLaMA-7b & 100\% \\
LLaMA-13b & 61\% \\
LLaMA-30b & N/A \\
OPT-13b & N/A \\
OPT-30b & N/A \\
BLOOM-560m & 96\% \\
BLOOM-3b & 100\%\\
Pythia-70m & 69\% \\
Pythia-410m & 37\% \\
Pythia-12b & 64\% \\
\hline
Avg. & 75\% \\
\hline
\end{tabular}
\caption{The gap closed in validation perplexity after fine-tuning between QLoRA 4-bit and QLoRA 8-bit quantization by QuAILoRA. The result for LLaMA 30b is omitted because the 8-bit model underperforms the 4-bit model. The results for OPT 13b and 30b are omitted because we were not able to generate results for 8-bit quantization.}
\label{table:perplexity_gap_closed}
\end{subtable}
\end{table*}

%% file: fig/downstream_tasks_train_split_summarized.tex
\begin{table*}[!t]
\caption{Downstream Task Results}
\label{table:downstream_task_results_summarized}
\begin{subtable}{0.45 \linewidth}
\centering
\begin{tabular}{|cccc|}
\hline
Model & Bits & Method & Avg. \\
\hline
LLaMA-7b & 4 & QLoRA & 62.1 \\
LLaMA-7b & 4 & Ours & \bf 62.8 \\
\hline
LLaMA-7b & 8 & QLoRA & 63.0 \\
LLaMA-7b & 8 & Ours & \textbf{63.1} \\
\hline
LLaMA-13b & 4 & QLoRA & {65.4} \\
LLaMA-13b  & 4 & Ours & \textbf{65.8} \\
\hline
LLaMA-13b & 8 & QLoRA & \textbf{65.8} \\
LLaMA-13b & 8 & Ours & {65.7} \\
\hline
\end{tabular}
\caption{Downstream task accuracy averaged across 7 downstream tasks for LLaMA models fine-tuned on Alpaca.}
\label{table:downstream_tasks_alpaca_summarized}
\end{subtable}
\quad
\begin{subtable}{0.45 \linewidth}
\centering
\begin{tabular}{|cc|}
\hline
Model & Gap Closed \\
\hline
LLaMA-7b & 74\% \\
LLaMA-13b & 100\% \\
\hline
\end{tabular}
\caption{The gap in average accuracy between 4-bit and 8-bit QLoRA models closed by QuAILoRA for models fine-tuned on Alpaca.}
\label{table:downstream_tasks_alpaca_gap_closed}
\end{subtable}

\begin{subtable}{0.45 \linewidth}
\centering
\begin{tabular}{|cccc|}
\hline
 Model & Bits & Method & Avg. \\
\hline
LLaMA-7b & 4 & QLoRA & 63.2 \\
LLaMA-7b & 4 & Ours & \textbf{63.9} \\
\hline
LLaMA-7b & 8 & QLoRA & {63.8} \\
LLaMA-7b & 8 & Ours & \textbf{63.9} \\
\hline
LLaMA-13b & 4 & QLoRA & 66.7 \\
LLaMA-13b & 4 & Ours & \textbf{67.0} \\
\hline
LLaMA-13b & 8 & QLoRA & \textbf{67.2} \\
LLaMA-13b & 8 & Ours & 67.1 \\
\hline
\end{tabular}
\caption{Downstream task accuracy averaged across 7 downstream tasks for LLaMA models fine-tuned on SlimOrca.}
\label{table:downstream_tasks_slimorca_summarized}
\end{subtable}
\quad
\begin{subtable}{0.45 \linewidth}
\centering
\begin{tabular}{|cc|}
\hline
Model & Gap Closed \\
\hline
LLaMA-7b & 89\% \\
LLaMA-13b & 84\% \\
\hline
\end{tabular}
\caption{The gap in average accuracy between 4-bit and 8-bit QLoRA models closed by QuAILoRA for models fine-tuned on SlimOrca.}
\label{table:downstream_tasks_slimorca_gap_closed}
\end{subtable}
\end{table*}

%% file: fig/lora_rank_fig.tex
\begin{figure*}
\centering
\includegraphics{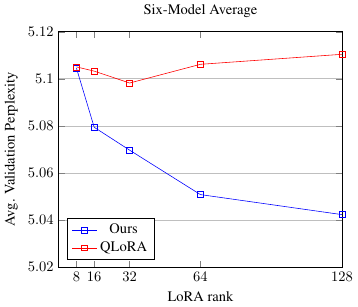}
\includegraphics{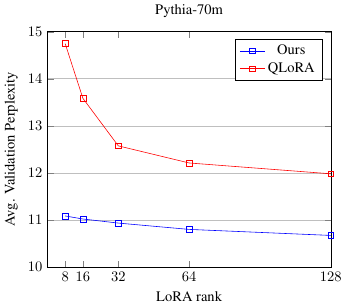}

\caption{Effect of LoRA rank on validation perplexity after fine-tuning 4-bit models, averaged across six 4-bit LLMs and 4 causal language modeling tasks. Increasing the LoRA rank results in a continual decrease in average validation perplexity when initializing with our method, albeit with diminishing returns. In contrast, increasing the LoRA rank does not significantly affect performance when using the baseline initialization. We plot results for Pythia-70m separately (not included in the six-model average) as this was the only baseline to show a strong decrease in validation perplexity with increasing LoRA rank.}
\label{fig:lora_rank}
\end{figure*}

%% file: fig/convergence_plot.tex
\begin{figure*}
\centering
\includegraphics{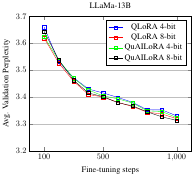}
\includegraphics{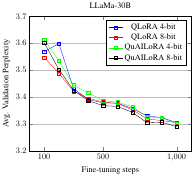}
\includegraphics{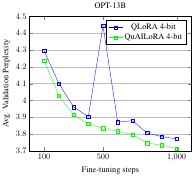}

\includegraphics{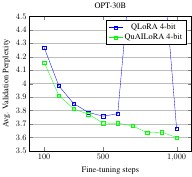}
\includegraphics{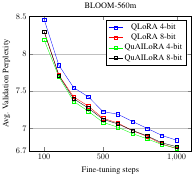}
\includegraphics{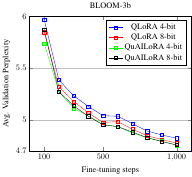}

\includegraphics{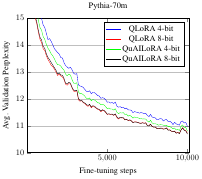}
\includegraphics{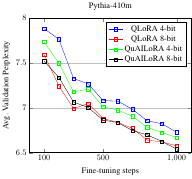}
\includegraphics{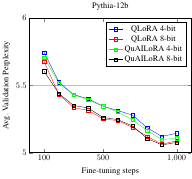}
\includegraphics{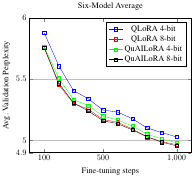}

\caption{Fine-tuning convergence for each of our 9 base models. We also include a six-model average convergence curve that excludes OPT-13b and OPT-30b (due to the perplexity spikes in the middle of fine-tuning) as well as Pythia-70m (which we fine-tune for 10 times as many steps as the other models).}
\label{fig:convergence_plots}
\end{figure*}

%% file: conclusion.tex
In this paper we introduced QuAILoRA, a method for increasing the performance of QLoRA without additional memory cost during fine-tuning by initializing the LoRA matrices to decrease quantization error. To find such an initialization, we optimized a calibrated quantization objective using alternating optimization, solving a small rank-$r$ linear system in each step. In our experiments, we demonstrated that our LoRA initialization can moderately improve the performance of QLoRA when the impact of quantization errors is significant. Furthermore, we found that our results are robust to the choice of the calibration dataset. 

%% file: limitations.tex
In our experiments, we showed that our method provides a small positive advantage over the baseline QLoRA when quantization error significantly impacts performance. However, quantization error appears to have less of a negative impact for the larger models, including the 13b parameter models we experimented with, and our method may provide only a small or no statistically significant advantage in such cases. We also only present experiments on models up to size 13b on downstream tasks, whereas experiments with models of size 30b and 70b are also common in the literature.

We only explore the affects of our method when quantizing to 4-bit or 8-bit. However, more recent work has also explored quantizing to 3, 2, or 1 bits. It is possible that our method provides a more significant positive advantage in these scenarios where the quantization error is expected to be larger, but we do not present experimental results for these scenarios.


%% file: appendix.tex
\input{fig/full_perplexity_table}

\newpage

\input{fig/downstream_tasks_train_split_full}

%% file: fig/full_perplexity_table.tex
\begin{table}[H]
\centering
\caption{Validation perplexity of QLoRA and QuAILoRA models.}
\begin{tabular}{|cccccccc|}
\hline
Model & Bits & Method & Alpaca & Chip2 & Self-Instruct & HH-RLHF & Avg. \\
\hline
LLaMA-7b & 4 & QLoRA & $3.69$ & $\bf 3.44$ & $\bf 2.08$ & $4.84$ & $3.51$ \\
LLaMA-7b & 4 & Ours & $\bf 3.65$ & $\bf 3.44$ & $2.09$ & $\bf 4.79$ & $\bf 3.49$ \\
\hline
LLaMA-7b & 8 & QLoRA & $\bf 3.65$ & $3.43$ & $2.08$ & $4.82$ & $3.49$ \\
LLaMA-7b & 8 & Ours & $\bf 3.65$ & $\bf 3.41$ & $\bf 2.06$ & $\bf 4.81$ & $\bf 3.48$ \\
\hline
LLaMA-13b & 4 & QLoRA & $3.45$ & $\bf 3.24$ & $\bf 2.11$ & $4.52$ & $3.33$ \\
LLaMA-13b & 4 & Ours & $\bf 3.44$ & $\bf 3.24$ & $\bf 2.11$ & $\bf 4.50$ & $\bf 3.32$ \\
\hline
LLaMA-13b & 8 & QLoRA & $3.44$ & $\bf 3.23$ & $\bf 2.09$ & $4.52$ & $3.32$ \\
LLaMA-13b & 8 & Ours & $\bf 3.43$ & $\bf 3.23$ & $\bf 2.09$ & $\bf 4.51$ & $\bf 3.31$ \\
\hline
LLaMA-30b & 4 & QLoRA & $\bf 3.42$ & $\bf 3.22$ & $\bf 2.10$ & $4.48$ & $\bf 3.30$ \\
LLaMA-30b & 4 & Ours & $\bf 3.42$ & $\bf 3.22$ & $\bf 2.10$ & $\bf 4.46$ & $\bf 3.30$ \\
\hline
LLaMA-30b & 8 & QLoRA & $3.44$ & $3.21$ & $2.10$ & $4.48$ & $3.31$ \\
LLaMA-30b & 8 & Ours & $\bf 3.42$ & $\bf 3.20$ & $\bf 2.08$ & $\bf 4.47$ & $\bf 2.29$ \\
\hline
OPT-13b & 4 & QLoRA & $3.89$ & $3.68$ & $2.25$ & $5.27$ & $3.77$ \\
OPT-13b & 4 & Ours & $\bf 3.82$ & $\bf 3.59$ & $\bf 2.21$ & $\bf 5.24$ & $\bf 3.71$ \\
\hline
OPT-30b & 4 & QLoRA & $3.78$ & $3.57$ & $2.14$ & $5.15$ & $3.66$ \\
OPT-30b & 4 & Ours & $\bf 3.68$ & $\bf 3.48$ & $\bf 2.13$ & $\bf 5.11$ & $\bf 3.60$ \\
\hline
BLOOM-560m & 4 & QLoRA & $6.85$ & $6.40$ & $3.67$ & $10.46$ & $6.84$ \\
BLOOM-560m & 4 & Ours & $\bf 6.73$ & $\bf 6.27$ & $\bf 3.60$ & $\bf 10.34$ & $\bf 6.73$ \\
\hline
BLOOM-560m & 8 & QLoRA & $\bf 6.70$ & $\bf 6.31$ & $\bf 3.60$ & $\bf 10.31$ & $\bf 6.73$ \\
BLOOM-560m & 8 & Ours & $6.71$ & $6.36$ & $3.63$ & $10.33$ & $6.76$ \\
\hline 
BLOOM-3b & 4 & QLoRA & $4.71$ & $4.44$ & $2.63$ & $7.51$ & $4.82$ \\
BLOOM-3b & 4 & Ours & $\bf 4.64$ & $\bf 4.35$ & $\bf 2.59$ & $\bf 7.45$ & $\bf 4.75$ \\
\hline
BLOOM-3b & 8 & QLoRA & $4.66$ & $4.39$ & $2.60$ & $7.45$ & $4.78$ \\
BLOOM-3b & 8 & Ours & $\bf 4.63$ & $\bf 4.36$ & $\bf 2.59$ & $\bf 7.44$ & $\bf 4.76$ \\
\hline
Pythia-70m & 4 & QLoRA & $13.55$ & $11.08$ & $6.26$ & $13.03$ & $10.98$ \\
Pythia-70m & 4 & Ours & $\bf 13.39$ & $\bf 10.86$ & $\bf 6.08$ & $\bf 12.87$ & $\bf 10.80$ \\
\hline
Pythia-70m & 8 & QLoRA & $13.18$ & $\bf 10.73$ & $\bf 5.97$ & $13.00$ & $10.72$ \\
Pythia-70m & 8 & Ours & $\bf 13.13$ & $10.74$ & $6.00$ & $\bf 12.90$ & $\bf 10.69$ \\
\hline
Pythia-410m & 4 & QLoRA & $7.61$ & $6.57$ & $4.18$ & $8.55$ & $6.73$ \\
Pythia-410m & 4 & Ours & $\bf 7.57$ & $\bf 6.52$ & $\bf 4.09$ & $\bf 8.50$ & $\bf 6.67$ \\
\hline
Pythia-410m & 8 & QLoRA & $7.42$ & $6.40$ & $4.12$ & $\bf 8.36$ & $6.57$ \\
Pythia-410m & 8 & Ours & $\bf 7.35$ & $\bf 6.38$ & $\bf 4.06$ & $\bf 8.36$ & $\bf 6.54$ \\
\hline
Pythia-12b & 4 & QLoRA & $5.93$ & $5.06$ & $3.08$ & $6.50$ & $5.14$ \\
Pythia-12b & 4 & Ours & $\bf 5.90$ & $\bf 5.00$ & $\bf 3.03$ & $\bf 6.50$ & $\bf 5.11$ \\
\hline
Pythia-12b & 8 & QLoRA & $5.83$ & $5.00$ & $3.04$ & $6.48$ & $5.09$\\
Pythia-12b & 8 & Ours & $\bf 5.81$ & $\bf 4.99$ & $\bf 3.03$ & $\bf 6.47$ & $\bf 5.08$ \\
\hline
\end{tabular}
\label{table:full_perplexity_table}
\end{table}

%% file: fig/downstream_tasks_train_split_full.tex
\begin{table}[H]
\centering
\begin{subtable}{\textwidth}
\centering
\begin{tabular}{|c|c|c|c|c|c|c|c|c|c|}
\hline
Method & Model & Arc-C & Arc-E & BoolQ & HS & OBQA & PIQA & WinoG & avg. \\
\hline
QLoRA & 7b 4-bit & 41.8\tiny$\pm$1.5 & 71.3\tiny$\pm$1.0 & 73.9\tiny$\pm$0.5 & 55.7\tiny$\pm$0.2 & 31.5\tiny$\pm$0.7 & 77.6\tiny$\pm$0.3 & 82.8\tiny$\pm$0.2 & 62.1 \\
Ours & 7b 4-bit & \textbf{42.2}\tiny$\pm$1.5 & \textbf{73.1}\tiny$\pm${0.9} & \textbf{75.7}\tiny$\pm${0.4} & \textbf{55.8}\tiny$\pm$0.2 & \textbf{31.9}\tiny$\pm$0.7 & \textbf{77.9}\tiny$\pm$0.3 & \textbf{82.9}\tiny$\pm$0.2 & \bf 62.8 \\
\hline
QLoRA & 7b 8-bit & \textbf{43.3}\tiny$\pm$1.5 & 72.3\tiny$\pm$0.9 & \textbf{76.8}\tiny$\pm$0.4 & 55.9\tiny$\pm$0.2 & 32.1\tiny$\pm$0.7 & \textbf{77.8}\tiny$\pm$0.3 & \textbf{83.0}\tiny$\pm$0.2 & 63.0 \\
Ours & 7b 8-bit & 42.8\tiny$\pm$1.5 & \textbf{73.3}\tiny$\pm$0.9 & 76.2\tiny$\pm$0.4 & \textbf{56.4}\tiny$\pm${0.2} & \textbf{32.5}\tiny$\pm$0.7 & \textbf{77.8}\tiny$\pm$0.3 & \textbf{83.0}\tiny$\pm$0.2 & \textbf{63.1} \\
\hline
QLoRA & 13b 4-bit & 44.6\tiny$\pm$1.5 & 74.3\tiny$\pm$0.9 & 82.6\tiny$\pm$0.4 & \textbf{59.2}\tiny$\pm${0.2} & {33.0}\tiny$\pm$0.7 & {79.4}\tiny$\pm$0.3 & {84.9}\tiny$\pm$0.2 & {65.4} \\
Ours & 13b 4-bit & \textbf{46.6}\tiny$\pm$1.5 & \textbf{74.5}\tiny$\pm$0.9 & \textbf{83.0}\tiny$\pm$0.4 & 58.5\tiny$\pm$0.2 & \textbf{33.2}\tiny$\pm$0.7 & \textbf{79.5}\tiny$\pm$0.3 & \textbf{85.2}\tiny$\pm$0.2 & \textbf{65.8} \\
\hline
QLoRA & 13b 8-bit & \textbf{45.8}\tiny$\pm$1.5 & \textbf{74.7}\tiny$\pm$0.9 & {82.6}\tiny$\pm$0.4 & \textbf{58.9}\tiny$\pm$0.2 & \textbf{33.5}\tiny$\pm$0.7 & \textbf{79.5}\tiny$\pm$0.3 & {85.4}\tiny$\pm$0.2 & \textbf{65.8} \\
Ours & 13b 8-bit & 45.6\tiny$\pm$1.5 & 74.5\tiny$\pm$0.9 & \textbf{82.8}\tiny$\pm$0.4 & {58.8}\tiny$\pm$0.2 & 33.3\tiny$\pm$0.7 & \textbf{79.5}\tiny$\pm$0.3 & \textbf{85.6}\tiny$\pm$0.2 & {65.7} \\
\hline
\end{tabular}
\caption{Downstream task performance for LLaMA models fine-tuned on Alpaca.}
\end{subtable}

\begin{subtable}{\textwidth}
\centering
\begin{tabular}{|c|c|c|c|c|c|c|c|c|c|}
\hline
 Method & Model & Arc-C & Arc-E & BoolQ & HS & OBQA & PIQA & WinoG & avg. \\
\hline
QLoRA & 7b 4-bit & 41.6\tiny$\pm$1.5 & 71.8\tiny$\pm$0.9 & 80.2\tiny$\pm$0.4 & 57.2\tiny$\pm$0.2 & {31.5}\tiny$\pm$0.7 & 77.1\tiny$\pm$0.3 & 83.2\tiny$\pm$0.2 & 63.2 \\
Ours & 7b 4-bit & \textbf{42.4}\tiny$\pm$1.5 & \textbf{73.3}\tiny$\pm$0.9 & \textbf{80.9}\tiny$\pm$0.4 & \textbf{57.6}\tiny$\pm$0.2 & \textbf{31.9}\tiny$\pm$0.7 & \textbf{77.4}\tiny$\pm$0.3 & \textbf{83.7}\tiny$\pm${0.2} & \textbf{63.9}  \\
\hline
QLoRA & 7b 8-bit & \textbf{41.8}\tiny$\pm$1.5 & 72.8\tiny$\pm$0.9 & 81.2\tiny$\pm$0.4 & 57.9\tiny$\pm$0.2 & \textbf{32.0}\tiny$\pm$0.7 & \textbf{77.6}\tiny$\pm$0.3 & 83.5\tiny$\pm$0.2 & {63.8} \\
Ours & 7b 8-bit & 41.6\tiny$\pm$1.5 & \textbf{73.0}\tiny$\pm$0.9 & \textbf{81.5}\tiny$\pm$0.4 & \textbf{58.1}\tiny$\pm$0.2 & 31.9\tiny$\pm$0.7 & {77.5}\tiny$\pm$0.3 & \textbf{83.7}\tiny$\pm$0.2 & \textbf{63.9} \\
\hline
QLoRA & 13b 4-bit & {48.0}\tiny$\pm$1.5 & 77.3\tiny$\pm$0.9 & \textbf{84.1}\tiny$\pm$0.4 & 59.8\tiny$\pm$0.2 & 33.4\tiny$\pm$0.7 & {79.4}\tiny$\pm$0.3 & 85.2\tiny$\pm$0.2 & {66.7} \\
Ours & 13b 4-bit & \textbf{48.3}\tiny$\pm$1.5 & \textbf{77.8}\tiny$\pm$0.9 & 83.9\tiny$\pm$0.4 & \textbf{59.9}\tiny$\pm$0.2 & \textbf{34.0}\tiny$\pm$0.7 & \textbf{79.6}\tiny$\pm$0.3 & \textbf{85.6}\tiny$\pm$0.2 & \textbf{67.0} \\
\hline
QLoRA & 13b 8-bit & \textbf{48.8}\tiny$\pm$1.5 & 78.1\tiny$\pm$0.9 & \textbf{84.6}\tiny$\pm$0.4 & 60.1\tiny$\pm$0.2 & \textbf{34.0}\tiny$\pm$0.7 & \textbf{79.5}\tiny$\pm$0.3 & \textbf{85.5}\tiny$\pm$0.2 & \textbf{67.2} \\
Ours & 13b 8-bit & \textbf{48.8}\tiny$\pm$1.5 & \textbf{78.3}\tiny$\pm$0.9 & 83.5\tiny$\pm$0.4 & \textbf{60.2}\tiny$\pm$0.2 & \textbf{34.0}\tiny$\pm$0.7 & \textbf{79.5}\tiny$\pm$0.3 & \textbf{85.5}\tiny$\pm$0.2 & 67.1 \\
\hline
\end{tabular}
\caption{Downstream task performance for LLaMA models fine-tuned on SlimOrca.}
\end{subtable}
\caption{Downstream task accuracy of QLoRA and QuAILoRA models. Error bars reported are one standard error.}
\label{table:downstream_tasks_train_split_full}
\end{table}